\documentclass{article}

\usepackage{packages}

\begin{document}



\def\TITLE{Facial Age Estimation Using Convolutional Neural Networks}

\def\GROUP{Group 8}

\def\AUTHORS{
    Adrian Kjærran (adriankj) \\
    Erling Stray Bugge (erlinsb) \\
    Christian Bakke Vennerød (christbv) \\
}


\begin{titlepage}

\vbox{ }

\vbox{ }

\begin{center}
\includegraphics[width=0.40\textwidth]{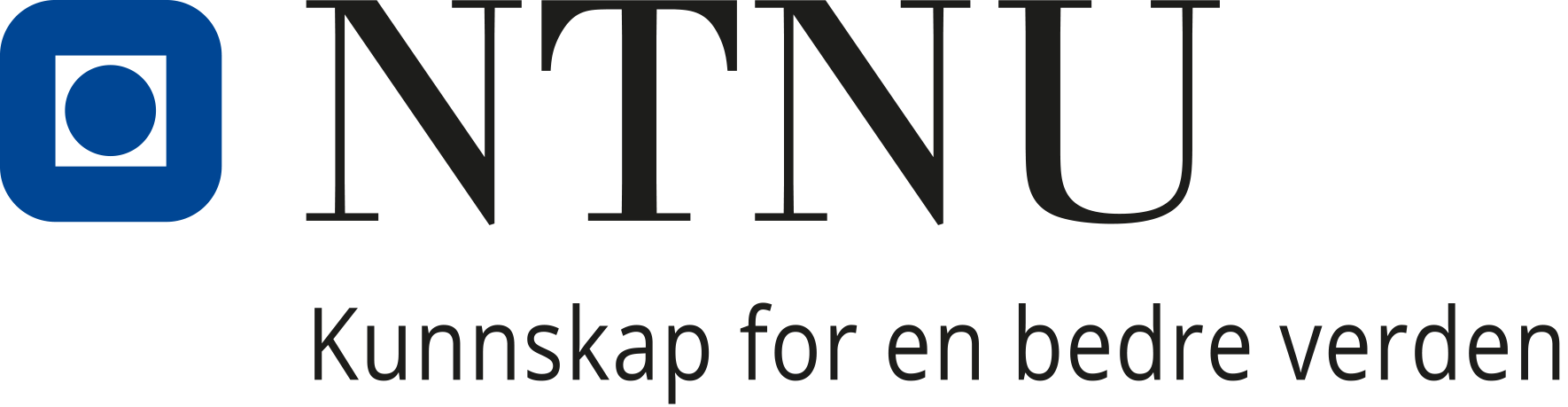}\\[1cm]
\textsc{\LARGE Department of Computer Science}\\[1.0cm]
\textsc{\Large TDT4173 - Final project}\\[0.5cm]

\def\checktitle{Paper Title}
\ifx\TITLE\checktitle
    \todo[inline]{
        Remember to fill in paper title, group name, and members in "title.tex"
    }
\fi

\vbox{ }
\HRule \\[0.4cm]
{ \huge \bfseries \TITLE}\\[0.4cm]
\HRule \\[1.5cm]
\large

\emph{Group:}\\
\GROUP

\emph{Authors:}\\
\AUTHORS

\vfill
{\large \today}
\end{center}
\end{titlepage}






\frontmatter

\begin{abstract}

\noindent This paper is a part of the final project in the course TDT4173, Machine Learning, at NTNU. In this paper, a deep convolutional neural network with five convolutional layers and three fully-connected layers is presented to estimate individuals' ages from images. The model is in its entirety trained from scratch, where a combination of three different datasets is used as training data. These datasets are the APPA dataset, UTK dataset, and IMDB dataset. The images were preprocessed using a face-recognition software. Our model is evaluated on both a held-out test set, and the Adience benchmark. On the test set, our model achieves a categorical accuracy of 0.52. On the Adience benchmark, our model proves inferior compared with another model, with an exact accuracy of 0.30, and an one-off accuracy of 0.46. Furthermore, a script was created, allowing users to estimate their age directly on their web camera. This script and all other code is located in the GitHub repository:  \href{https://github.com/christianbv/AgeNet}{AgeNet}.

\end{abstract}
\clearpage

\tableofcontents

\listoffigures
\listoftables

\mainmatter

\titlespacing\section{0pt}{8pt plus 2pt minus 2pt}{2pt plus 5pt minus 5pt}
\section{Introduction} 
\label{sec:introduction}

There are numerous situations where it is relevant to estimate the age of a person. Either it is for customer profiling, security measures for digital signatures, or intelligent video surveillance, a model capable of solving this task would be of immense value to society \citep{dornaika_et_al}. 

Due to the inherent features of a person's face, precisely estimating the age based on an image is difficult. Facial aging variation is complicated and specific to a given individual as well as many external factors such as lifestyle and climate \citep{sai_et_al}. Two individuals of the same real age could therefore appear to be of two vastly different ages. Furthermore, an ordinal relationship and correlation between the age labels exist. The age of 40 is closer to 35 than 10. This makes age estimation a more demanding task compared to a problem where there is no correlation between the classes \citep{chao_et_al}. 

This paper seeks to address the problem of estimating an individual's age by performing supervised age range estimation. This will be done by multi-class classification. Since the ImageNet competition in 2012 and the great results of the deep CNN model developed by \cite{alex_net}, CNNs have been widely used on large scale image classification tasks \citep{gu_et_al}. Today, convolutional neural networks have achieved outstanding results when it comes to facial recognition \citep{deepface}. Therefore, it is evident that deep learning and convolutional neural networks are a relevant tool for this task.

The data used to train the model was a combination of three different image datasets. These were the APPA dataset, with 7 200 images, the UTK dataset, with 22 000 images, and the IMDB dataset, with 37 000 images. The performance of our approach is evaluated on the Adience dataset, a widely used benchmark on multi-class age classification \citep{adience_authors}.

\section{Related Work} 
\label{sec:related_work}
The current research on age estimation from images can be divided into two sub-categories based on how the estimations are made. The problem can either be tackled as a multi-class classification problem, where the age spectre is converted into age bins, or as a regression problem, where the exact age is predicted \citep{dornaika_et_al}. 

One example of this is that of \cite{IMDB_dataset}, who used transfer learning and deep CNNs for multi-class classification. They achieved great results by fine-tuning the VGG-16 model on the IMDB-WIKI dataset. 

\cite{chao_et_al} used a regression approach by creating a k-Nearest Neighbour Support Vector Regression model for determining the numerical age of a person from images. Another regression approach was that of \cite{dornaika_et_al}, who used $l1$-norm error and an adaptive loss function with CNNs to obtain great results.

There are also related work which combines the two sub-categories mentioned \citep{hierarchical}. These are called hierarchical approaches, where both classification and regression are used to obtain a prediction within yearly granularity \citep{dornaika_et_al}. One such hierarchical approach was that of \cite{pontes_et_al}. They divided facial features into local and global features when preprocessing the images. The features were then concatenated and used in a Support Vector Machine (SVM) for age group classification. Lastly, the classifications from the SVM were used as input in a Support Vector Regression (SVR) model to obtain a numerical age estimation. 

In this paper, we have chosen a multi-class classification approach. However, we have currently only found papers where the classes had gaps between the bins, e.g., $\{0-2, 4-6, 8-13, \dots\}$. One advantage of this bin-format is that by only considering images within these ages, the data features might be more discriminative. This allows a model to greater extent learn the most significant facial features of different age groups. On the other hand, not all ages are represented in the data, thereby limiting the practical applicability of the model. We have, therefore, used continuous bins in this paper, i.e., bins with no gap between them. As we use multiple datasets and continous bins, we believe our solution can \emph{contribute} to the existing research on the matter. 

\titlespacing\section{0pt}{4pt plus 2pt minus 2pt}{2pt plus 5pt minus 5pt}
\titlespacing\subsection{0pt}{4pt plus 2pt minus 2pt}{1pt plus 5pt minus 5pt}
\section{Data}
\label{sec:data}
The data used for training our model and producing the results are from three different datasets, all of which solely contain images of people labeled by their age. The three datasets are the APPA dataset, the UTK dataset, and the IMDB dataset.  

\subsection{Data characteristics}
\begin{figure}[H]
\begin{minipage}{0.5\textwidth}
\centering
\captionsetup{justification=centering}
\includegraphics[width = 0.9\textwidth]{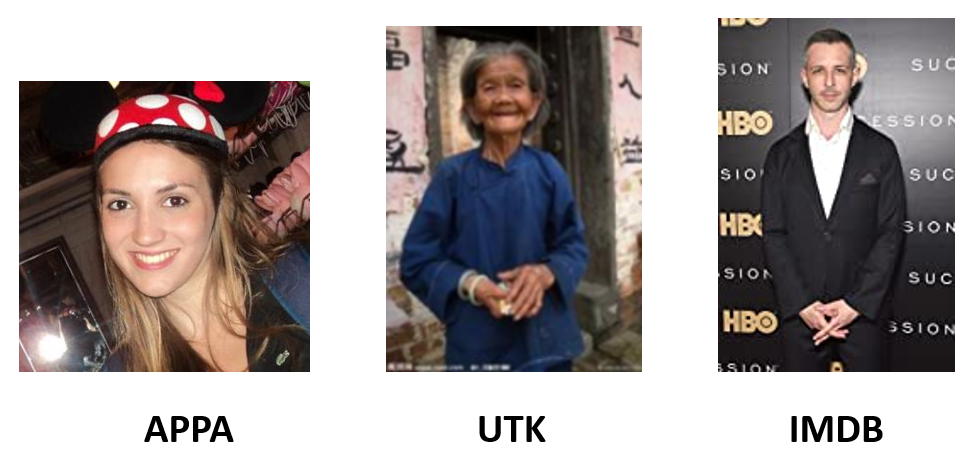}
\caption{Sample images.}
\label{fig:sample_images_before_processing}
\end{minipage}
\begin{minipage}{0.5\textwidth}
\centering
\begin{tabular}{c|ccc}
\textbf{Dataset}             & \textbf{Images} & \textbf{Mean} & \textbf{STD} \\ \hline
\textit{\textbf{Aggregated}} & 66 358       & 36,9              & 16.5           \\
\textit{\textbf{APPA}}       & 7 202          & 31.3              & 17.1          \\
\textit{\textbf{UTK}}        & 22 000         & 33.2              & 20.2         \\
\textit{\textbf{IMDB}}  & 37 156          & 40.2              & 12.7        
\end{tabular}
\captionof{table}{Summary of data statistics.}
\label{tab:summary_data_statistics_2}
\end{minipage}
\end{figure}

The APPA dataset consists of approximately 7 200 images of people. These images have a wide variety of angles, context, and lighting features. Compared to our two other datasets, the APPA dataset has the most even distribution of ages, which can be seen in Table \ref{tab:summary_data_statistics_2}.

The UTK dataset consists of approximately 22 000 in-the-wild pictures. These are taken in various conditions and angles. The dataset has a mean age of 33.2 but also contains many pictures of younger people. These images are important in order to ensure that the model learns the dependencies found in the younger age bins. 

The IMDB dataset originally contains 450 000 images, but out of these, only 37 000 were used. These 37 000 images were sampled due to two reasons. Firstly, with a mean age of 40.2, the dataset had a low number of instances for the younger age groups. As the other datasets also had the same problem, but to a lower extent, reducing the size of the IMDB-dataset made the younger age groups better represented in the aggregated dataset.
Secondly, most of the images in the dataset included actors with makeup and in perfect lighting conditions, making the data less representative for the normal population.

\subsection{Data Preprocessing}

\autoref{fig:sample_images_before_processing} shows sample images taken from each of the datasets. These pictures highlight some of the variances in distance, lighting, angles, and resolutions found in the data. The differences between images make it difficult for the model to find and learn patterns. Therefore, the data needs preprocessing before training the model. The preprocessing pipeline is as follows:

\titlespacing\subsubsection{0pt}{8pt plus 2pt minus 2pt}{-5pt plus 0pt minus 0pt}
\subsubsection{Step 1: Detect the face}
\label{sec:face_detection_section}
The first step in the preprocessing pipeline was to find the pixel coordinates for each face in each image. This is important because we want to remove all background noise and only train the age estimating model to evaluate the faces in the images. To achieve this, each image was processed using the \emph{face\_recognition} module from \cite{face_recognition}, which is built on \emph{dlib's} state-of-the-art solution for face recognition \citep{dlib}. The model from \emph{dlib} uses a CNN and has an accuracy of 99.38\% in detecting faces on the \textit{Labelled Faces in the Wild}-benchmark \citep{LFW}.

\subsubsection{Step 2: Crop and reshape resolution}
After detecting a face, we add a 40\% margin to ensure that the whole face is captured. Using the coordinates from step 1, the pipeline then used the \emph{Pillow} module to crop the picture and change its resolution to fit a size of 3x256x256 (RGB with height and width of 256 pixels) \citep{PIL}. 

\subsubsection{Step 3: Grouping images into bins} 
The third step of the pipeline was to group images into bins, where each bin represented a distinct age-range. Different combinations were tried, with the goal of replicating typical aging periods where individuals are as similar as possible. For the lower ages in the data, few years were included in each bin to ensure that individuals in each bin are as similar as possible. The result was to use the following bins: $\{(0-2),(3-6),(7-12),(13-17),(18-22), (23-26), (27-33), (34-44),(45-59),(60+)\}$, which correspond to 10 different classes. 

\subsubsection{Step 4: Cleaning}
The fourth step was to validate the labels of our images. Several images in the IMDB dataset were clearly mislabelled. Inaccuracies in the IMDB data was also found by \citep{mislabellings}. Removing wrongly labelled data is an important step in our pipeline, as the performance of the resulting models are highly dependant on the quality of the data. To handle this, a tentative age estimating CNN model was trained on the APPA and UTK datasets, and used to filter clearly mislabelled images from the IMDB dataset. An image was classified as mislabelled and thus removed if the model predicted less than a 40\% chance that the image belonged to the labelled bin or its two adjacent bins. 

\subsubsection{Step 5: Data Augmentation}
Finally, the data was augmented. This was done using TensorFlow's in-built methods. Data augmentation is a technique to increase the diversity of the training set, where the data is subject to random transformations. Three data transformation steps were implemented. These were random flip, i.e., horizontally flipping the image, random rotation, i.e., rotating the image a given number of degrees (helps the model become rotation invariant), and random zoom, i.e., randomly zooming in or out.

\subsection{Result}
\begin{figure}[H]
\begin{minipage}[b]{0.4\textwidth}
\centering
\captionsetup{justification=centering}
\includegraphics[width=\textwidth]{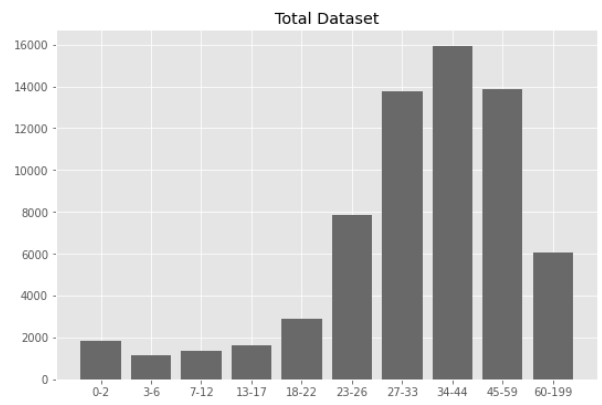}
\caption{Distribution of age bins.}{\emph{(X-axis:)} Age bins. \emph{(Y-axis):} Number of instances.}
\label{fig:data_distributions_2}
\end{minipage}
\begin{minipage}[b]{0.6\textwidth}
\centering
\captionsetup{justification=centering}
\includegraphics[width=\textwidth]{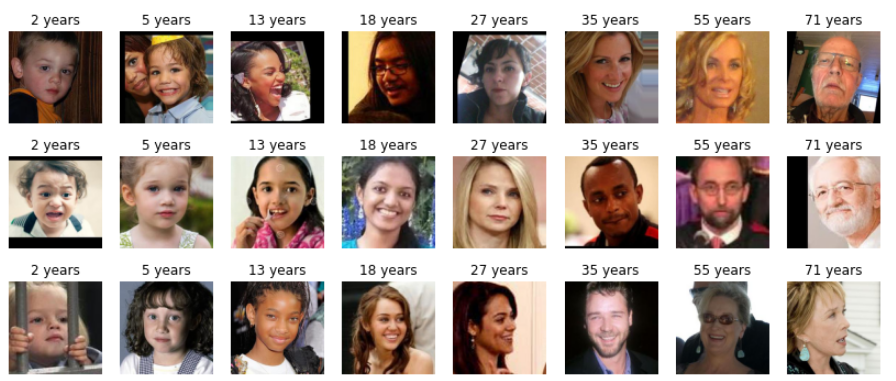}
\caption{Sample images after preprocessing.}{\emph{(Top):} APPA dataset. \emph{(Middle):} UTK dataset. \emph{(Bottom):} IMDB dataset.}
\label{fig:sample_images_2}
\end{minipage}
\end{figure}

Our preprocessing pipeline transformed the 66 000 images into a collection of centred and labelled pictures of faces, as shown in \autoref{fig:sample_images_2}. The distribution of labels can be seen in \autoref{fig:data_distributions_2}. From the figure, we observe an uneven distribution. Although we have fewer pictures in the younger ages, we believe with respect to the aging process that a model can more easily differentiate between the younger bins compared to older bins. Our processed dataset can be downloaded at \citep{ntnu_bucket}.

\subsection{Splitting strategy}
During training, the processed dataset was shuffled and divided into a training set (80\%), and a validation set (20\%). Each of the 91 tested models was trained from scratch on the training set and validated on the validation set. After the final model architecture and hyperparameters were found, a model was trained from scratch with 80 \% of the data as training data, and 20\% set a side for testing. The results of the model are further described in \autoref{sec:results}.

\titlespacing\section{0pt}{8pt plus 2pt minus 2pt}{5pt plus 5pt minus 5pt}
\titlespacing\subsection{0pt}{8pt plus 2pt minus 2pt}{5pt plus 5pt minus 5pt}
\newpage
\section{Methods}
\label{sec:methods}

To solve the problem of age estimation, we will be using a convolutional neural network (CNN). In this section, we will explain the choice of model, training objective, and describe the use of empirical results to obtain the optimal architecture of the CNN with respect to the data. Additionally, this section will also seek to clarify our use of third party implementations for solving this task.

\subsection{Overview of CNNs}
Convolutional neural networks (CNNs) is a class of neural networks especially designed to handle grid-like data, such as time series or images \citep{goodfellow_et_al}. Using the terminology from \cite{goodfellow_et_al}, a CNN typically consists of convolutional layers (affine convolutional operations, activation function and pooling) and fully-connected layers (FC). Each convolutional layer performs feature extraction, whereas the FC layers use the results from the convolutional layers to classify the images into a given label. 

\subsubsection{Why CNNs were used on this problem}
There are three main reasons why CNNs were chosen as the architecture for the problem in this paper.

Firstly, to build a model that can efficiently learn from image data, it is important that the model understands that pixels within close proximity most likely are related. When utilising traditional multilayer perceptrons (MLPs) on image data, these relations are lost, as the spatial structure of the data is removed when images are flattened into one-dimensional vectors and fed to the MLP \citep{goodfellow_et_al}. In contrast, CNNs can operate directly on two-dimensional images, allowing for the preservation of spatial structure in the data \citep{zhang2020dive}.

Secondly, in contrast to traditional FC networks, a CNN performs parameter sharing and has sparse connectivity, meaning that the number of floating point operations and memory requirements are significantly reduced - thus allowing the network to work on larger amounts of data on scarce computational resources \citep{goodfellow_et_al}. Despite being sparsely connected, the receptive field of CNNs increases with depth, allowing a unit in deeper layers to be indirectly connected to much, or even all, of the input space \citep{goodfellow_et_al}. 

Thirdly, due to the pooling layers, CNNs have the ability to deal with translation invariance, i.e, positional shifts of targets in the image, something that is crucial when dealing with facial images \citep{zhang2020dive}. As described in \autoref{sec:data}, the data was tilted or adjusted to increase the model's ability to generalize. The CNNs pooling layers make the model invariant to these artificial augmentations, as well as the translation invariances occurring in real images. 

\subsection{Training objective}
\label{sec:Training_objective}
During training, the weights in the fully-connected layers and the kernels in the convolutional layers are learned, while the hyperparameters such as kernel size, padding, and learning rate are all set before training. In order to make the model perform as good as possible, choosing the correct loss function, optimizer and evalation metric is of great importance. 

After preprocessing the data, we had a multi-class classification problem with 10 different classes, one class per age bin. Therefore, categorical cross-entropy was implemented as a loss function. Categorical cross-entropy weights the probabilities of a prediction when estimating the loss. The more certain the model predicts the right class, the lower the categorical cross-entropy loss. 

With $k$ classes, a training instance $i$ can be labelled as a one-hot encoding $\bm{y_{i}} = (y_{i,1}, \dots, y_{i,k})$, where $y_{i,j}$ is 1 if instance $i$ belongs to class $j$, and 0 otherwise. Multiplying this one-hot encoding with the softmax-output from the model, $p_{\theta}(y_{i,j} \mid \bm{x_i})$, 
gives the models probability for the correct label, as shown in \autoref{eqn:prediction}.

\begin{equation}
    \bm{\hat{y_i}} = p_\theta(\bm{y_i} \mid \bm{x_i}) = \sum_{j=1}^{k} y_{ij} p_{\theta}(y_{i,j} \mid \bm{x_i})
    \label{eqn:prediction}
\end{equation}
Based on that, for $n$ training instances, the categorical cross entropy loss can be defined by:
\begin{equation}
    CE(\bm{y}, \bm{\hat{y}}) = - \sum_{i=1}^{n} \bm{y_i} \log \bm{\hat{y_i}} = - \sum_{i=1}^{n} \sum_{j=1}^{k} y_{i,j} \log p_{\theta}(y_{i,j} \mid \bm{x_i})
\end{equation} 

In other words, our training objective is to minimize the categorical cross entropy loss.

\subsection{Optimizers}

The optimization algorithm decides how the weights in the fully-connected layers and kernels adjusts with respect to the loss-function. Finding the right combination of optimizer and hyperparameters are crucial for finding the best weights. In our project, three different optimizers were tested. These were SGD, RMSprop and Adam. Adam (Adaptive Moment Estimation) yielded the best results on both models. This corresponds with the findings of \cite{prillianti_et_al}, who found that Adam performed best on three different CNNs - a shallow net with only one convolutional layer, Le-net \citep{le_net}, and AlexNet \citep{alex_net}. For the interested reader an explanation of the Adam optimizer can be found in appendix \autoref{app:adam}.





\subsection{Third party implementations}
The group chose to work in Google Colaboratory (colab), an online Jupyter notebook instance which allows for fast deployment of code. Colab allows for free, limited usage of Graphical Processing Units (GPUs), which were of great benefit during the training of our model. Furthermore, colab was connected to Google Storage Buckets, enabling fast synchronization of data between the training instance and the data storage location. Our storage bucket is found in \cite{ntnu_bucket}.

\subsubsection{TensorFlow}
In colab, TensorFlow (TF) version 2.3.0 was used. TensorFlow is a high-level Python library, and allows for efficient deployment of machine learning models \citep{tensorflow}. An important feature of TensorFlow is its monitoring device, TensorBoard. TensorBoard is a GUI which makes tuning of hyperparameters and visualization of error metrics easier. \autoref{fig:tensorboard_tuning} shows a snapshot from TensorBoard during training of our model. 

\subsubsection{Face Detection}
A face detection library was used for detecting faces and cropping the images around the faces with a given margin. The library comes from \cite{dlib}, and uses a highly precise CNN with for detecting faces. This library was used both in colab for preprocessing images, as explained in \autoref{sec:face_detection_section}, and in a demonstration script for inferring ages through a web camera.

\subsection{Structure and Hyperparameter Tuning}
\begin{figure}[H]
    \centering
    \includegraphics[width=1\textwidth]{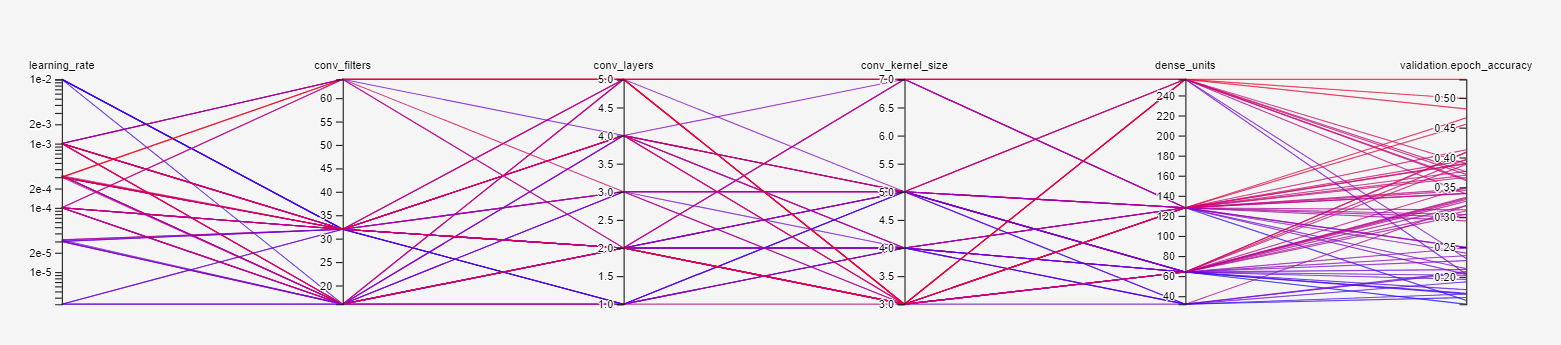}
    \caption{Tuning of hyperparameters in TensorBoard. From left to right: \emph{Learning rate}, \emph{conv\_filters}, \emph{conv\_layers}, \emph{conv\_kernel\_size}, \emph{dense\_units}, \emph{validation\_epoch\_accuracy}. }
    \label{fig:tensorboard_tuning}
\end{figure}

Through visualising the results in TensorBoard, several models with different structures and hyperparameters could be trained and analyzed in a structural manner. Starting with a wide range of hyperparameters and structures, and then sample uniformly from these sets, we gradually built an understanding of the best combinations. Hyperparameters which performed well, that is, had a high validation accuracy, were more densely sampled by repeating the process on a smaller search space with higher granularity.
\newpage
\autoref{fig:tensorboard_tuning} shows the tuning process. Each connected line from left to right represent one model. By choosing the models with highest validation accuracy (right-most axis/red color), and following their lines, the correct hyperparameters and model structure variables can be identified. In this case, we can observe that the learning rate (left-most column) probably should be between $3e^{-5}$ and $1e^{-2}$. 

\begin{figure}[H]
\centering
\begin{subfigure}{.5\textwidth}
  \centering
\includegraphics[width=\textwidth]{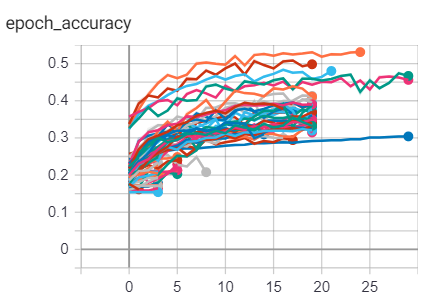}
    \caption{}
    \label{fig:accuracy_tensorboard}
\end{subfigure}%
\begin{subfigure}{.5\textwidth}
  \centering
 \includegraphics[width=\textwidth]{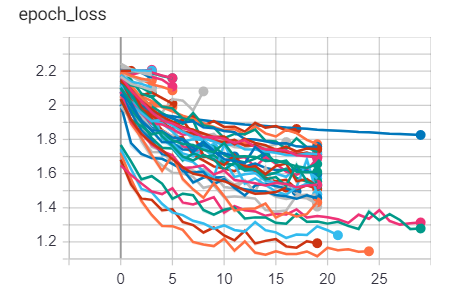}
    \caption{}
    \label{fig:loss_tensorboard}
\end{subfigure}
\caption{Validation accuracy and cross categorical loss per epoch, for a selected set of 91 trained models. Note that some of the models were only trained for 20 epochs.}
\label{fig:test}
\end{figure}

To avoid overfitting the training set, early stopping was implemented to ensure that a model stopped training if the error on the validation set did not improve over four epochs. \autoref{fig:accuracy_tensorboard} and \autoref{fig:loss_tensorboard} shows the validation accuracy and validation loss per epoch for a selected set of the 91 models that were trained, respectively. These plots were used in combination with \autoref{fig:tensorboard_tuning} for the selection of models. Ideally, we would want a model that has a steady increase in validation accuracy and a steady decrease in validation loss. In \autoref{fig:test}, the orange line shown in both plots, i.e., top of the left plot and bottom-most in the right plot, might be the best performing model in this subset. Furthermore, the results from each model was logged and saved. All of these logs are found in our GitHub repository, which is linked to in the appendix, section \ref{app:github_repo}.

\titlespacing\section{0pt}{8pt plus 2pt minus 2pt}{5pt plus 5pt minus 5pt}
\titlespacing\subsection{0pt}{4pt plus 2pt minus 2pt}{5pt plus 3pt minus 5pt}
\section{Results and Discussion} 
\label{sec:results}
This section will first describe our resulting model architecture, along with the chosen model structure, variables and hyperparameters. Secondly, the model will be evaluated in four different ways. After the result on each evaluation method is presented, a short discussion of the derived result will follow. 

\subsection{Final model architecture}

\begin{figure}[H]
    \centering
    \includegraphics[width=\textwidth]{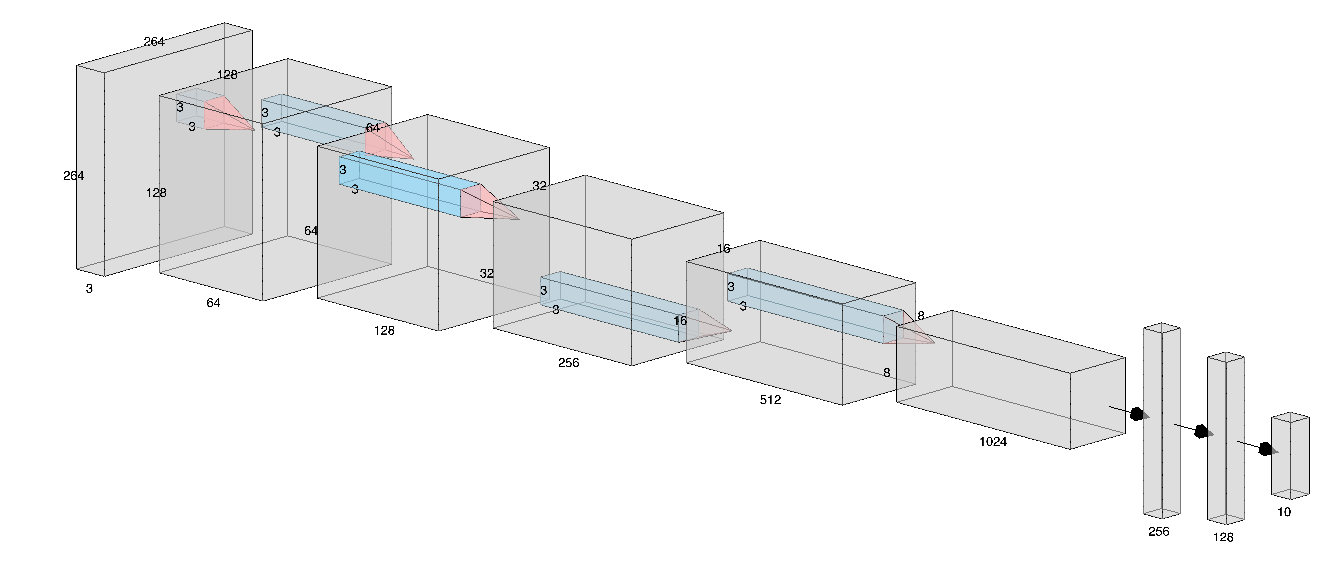}
    \source{Created with: http://alexlenail.me/NN-SVG/AlexNet.html}
    \caption[Resulting model architecture.]{Chosen model architecture. The 3 last grey boxes correspond to FC layers, whereas the blue rectangles correspond to 3x3 kernels in convolutional operations.}
    \label{fig:res_architecture}
\end{figure}

After having validated a total of 91 models, we ended up with the architecture shown in \autoref{fig:res_architecture}. Our approach is a CNN, which consists of five convolutional layers, where each convolutional layer also includes a Rectified Unit (ReLU) activation function layer, given by $max(0,x)$. The first convolutional layer applies 64 filters of size 3x3, with a stride of 2. For each of the next layers, the filter size is constant, but the number of filters is doubled. The pooling operation within each convolutional layer was max-pooling. Max pooling was chosen as it highlights textual features, unlike average pooling which averages the values.

After the convolutional layers, the model has three fully-connected layers, with 256, 128, and 10 neurons, respectively. These layers are responsible for learning dependencies and relations from the abstracted features extracted by the convolutional layers and convert these into predictions of the age. The final layer in the model is a softmax layer, which yields a ten-dimensional vector with probabilities between 0 and 1. Table \ref{tab:final_structure} and Table \ref{tab:final_hyperparameters} summarizes the model structure and chosen hyperparameters, respectively. Furthermore, the link to our final model is found in the appendix, section \ref{app:github_repo}.

\begin{figure}[H]
\begin{minipage}[t]{0.5\textwidth}
\centering
    \begin{tabular}[t]{c|c}
    \textbf{Structure}  & \textbf{Our model} \\ \hline
    \textbf{Conv layers} & 5 \\
    \textbf{Pooling layers} & Layer 1,2,3,4 \\  
    \textbf{Dropout layers} & Layer 1,2,3,4,5   \\
    \textbf{Conv filters}  & 64,128,256,512,1024    \\       
    \textbf{Kernel size} & 3x3      \\
    \textbf{} &  \\
    \textbf{Dense layers} & 3 \\
    \textbf{Dence neurons } & 256, 128, 10 \\
    \end{tabular}
    \captionof{table}{Summary of model structure.}
    \label{tab:final_structure}
\label{fig:foobar1}
\end{minipage}
\begin{minipage}[t]{0.5\textwidth}
\centering
    \begin{tabular}[t]{c|c}
    \textbf{Hyperparameters}  & \textbf{Our model} \\ \hline
    \textbf{Batch size}     & 32                  \\
    \textbf{Learning rate} & 0.0003                    \\
    \textbf{Dropout}        & 0.187                           \\
    \textbf{Optimizer} & Adam \\
    \textbf{Loss-function} & Cat. cross-entropy \\
    \end{tabular}
    \captionof{table}{Summary of model hyperparameters.}
    \label{tab:final_hyperparameters}
\end{minipage}
\end{figure}
\newpage
\subsection{Experiments and results}
\vspace{0.2cm}
The model was evaluated in four different ways. First, the feature maps were checked to ensure that the model has actually learned. Secondly, the performance of our model was evaluated on our test data. Thirdly, a confusion matrix was created to more thoroughly understand the results of our model. Finally, the performance of our model was evaluated on the Adience benchmark.  

\subsubsection{Visualization of results}
\begin{figure}[H]
    \centering
    \includegraphics[width=\textwidth]{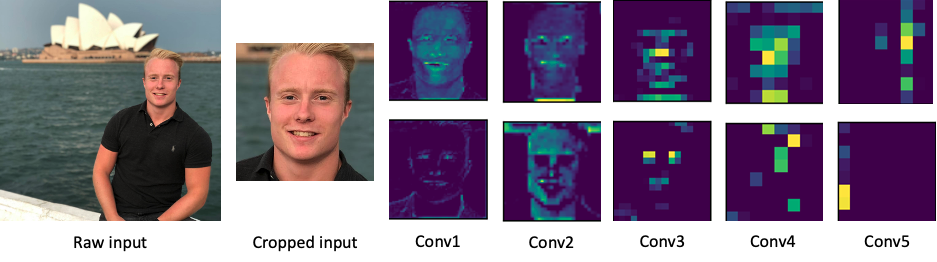}
    \caption{Visualization of feature maps in our convolutional network.}
    \label{fig:feature_maps}
\end{figure}

\autoref{fig:feature_maps} shows some key steps of our model when predicting the age of a person. The two first images show the preprocessing step, where the age is identified, and the image is cropped. The five columns of images to the right are some selected feature maps from each convolutional layer.

We observe that the model successfully filters out noise from the image and detects key features from the face of the subject. Furthermore, we observe that in the deeper convolutional layers, more complex patterns are inferred by the model. 

\vspace{0.2cm}

\subsubsection{Accuracy on test data}
The performance of our model on the training and test set is shown in \autoref{tab:summary_data_statistics}, where categorical accuracy was used as an evaluation metric. Note that the exact accuracy is used here and that the model performs best on the IMDB dataset, with $56\%$ accuracy. 


One reason for this could be explained by the fact that the IMDB images, in general, were of better quality and had better lighting conditions compared to the other in-the-wild datasets. Another reason could be that the larger portion of the data came from the IMDB dataset, thus making the model perform better on IMDB data. Finally, the preliminary model that was used to filter out mislabellings would probably remove difficult images and thus increase the performance on the IMDB dataset.


\begin{table}[H]
    \centering
    \large
\begin{tabular}{c|ccc}
\textbf{Dataset}             & \textbf{Train}   & \textbf{Test} \\ \hline
\textit{\textbf{Aggregated}} & 0.5875             & 0.520               \\
\textit{\textbf{APPA}}       & 0.462            & 0.358                   \\
\textit{\textbf{UTK}}        & 0.557           & 0.496                 \\
\textit{\textbf{IMDB}}  & 0.630          & 0.566                  
\end{tabular}
    \caption{Summary of the model's accuracy on training set and test set.}
   \label{tab:summary_data_statistics}

\end{table}

\subsubsection{Confusion matrix}

\begin{minipage}[t]{0.48\textwidth}
\vspace{0pt}
\autoref{fig:conf_matrix} shows the confusion matrix from testing our model on the test data. The x-axis is the predicted classes, and the y-axis is the ground truths. Observe that the model has a relatively high, exact accuracy for the age groups $(0-2), (3-6), (27-33), (45-59), (60+)$, and performs the worst on the age group $(23-26)$. 

\vspace{0.2cm}

These results correspond with our initial thoughts that the model will have difficulties with predicting some age groups due to perhaps few distinctive features. An interesting finding from this matrix is that the model overestimates the ages of individuals in the classes $(18-22)$ and $(23-26)$. One possible reason for this is the skewness in our data. As most data is in the range from $27$-$59$, the model could therefore be slightly biased towards overestimating the ages of people.

\end{minipage}
\begin{minipage}[t]{0.48\textwidth}
\vspace{-15pt}
\begin{figure}[H]
    \centering
    \includegraphics[width=\textwidth]{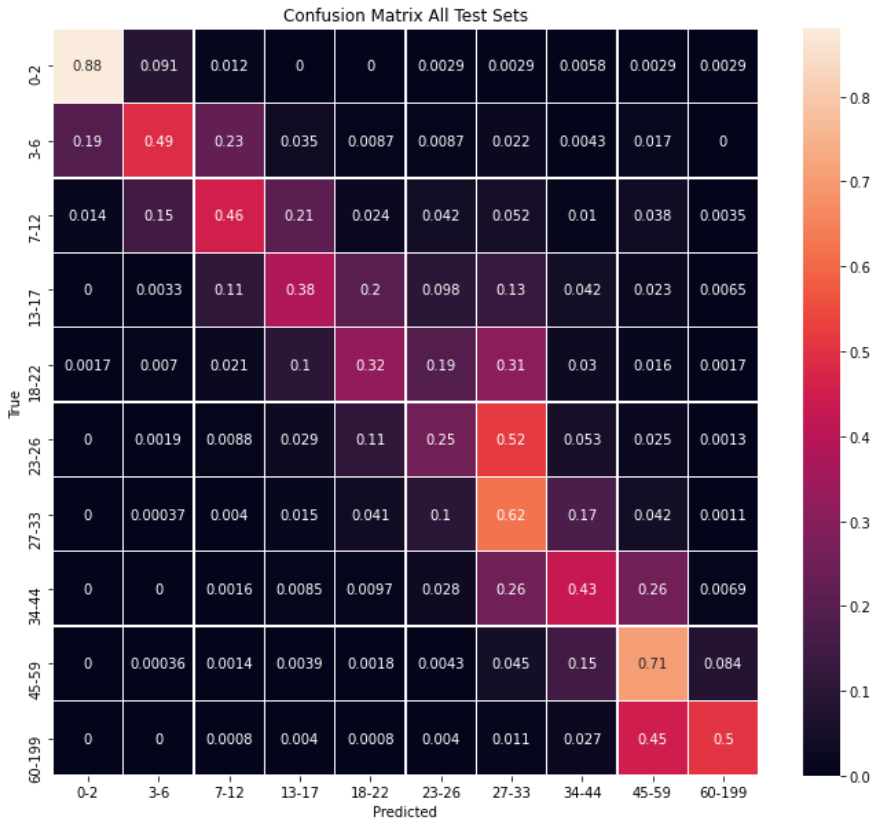}
    \caption{Confusion matrix showing the performance of the model for the various bins (classes).}
    \label{fig:conf_matrix}
    \hspace*{\fill}
\end{figure}
\end{minipage}



\subsubsection{Model evaluation on Adience benchmark}

Our model was also evaluated on the Adience benchmark, which is a popular benchmark for assessing the performance of age estimation models. Adience has the following classes: $(0-2),(4-6), (8-13), (15-20), (25-32), (38-43), (48-53), (60+)$. There are, however, two problems with the Adience benchmark: the age bins are different from the ones our model was trained on, and that there are gaps between each bin. To solve this, a hierarchical approach was implemented when evaluating the model on the Adience benchmark. The average of each bin, e.g. $42.5$ for the bin $(40-45)$, was multiplied with the softmax probability distributions from the output of our model. By doing this, a numerical yearly estimation of age is obtained. This was used to obtain two types of metrics. The first metric was exact accuracy, where we checked if the predicted age was within the label interval, and the second was one-off accuracy, where we checked if the predicted age was within the labeled bin or in the adjacent bins.

The results of our model evaluated on the Adience benchmark are shown in \autoref{tab:adience_results_v2}. Here, the performance of our model is compared to the original Adience paper by \cite{adience_authors}, which used a local binary pattern (LBP) model. From the table, it becomes clear that our model is inferior to that of \cite{adience_authors} when tested on the Adience dataset. 

We believe the result underestimates the performance of our model for multiple reasons. Firstly, the model was not trained to distinguish between the ages that were used in the Adience dataset. Secondly, our model was trained on continuous bins, which is not the case in the Adience dataset. Thirdly, the regression step that was implemented probably lead to more misclassifications. If the model predicts a young person to even have the slightest probability of being old, this would greatly increase the numerical age of this young person.

\begin{table}[H]
\centering
\begin{tabular}{c|ccc}
\textbf{Metric}           & \textbf{LBP} & \textbf{Our Model} & \textbf{Difference} \\ \hline
\textit{Exact accuracy}   & 0.414        & 0.304              & -0.110               \\
\textit{One-off accuracy} & 0.782        & 0.463              & -0.319              
\end{tabular}
\caption{Evaluation metrics for our model compared to the original Adience model \citep{adience_authors}. Number of images predicted were 11 964. }
\label{tab:adience_results_v2}
\end{table}

\section{Conclusion} 
\label{sec:conclusion}

In this paper, our approach to solving the age estimation problem is presented. By using data from three different datasets, an aggregated dataset with 60 000 facial images was created. Each image was preprocessed using an external face-recognition module and cropped around the face. These images were used to train our deep convolutional neural network. The model was trained from scratch, and has five convolution layers and three fully-connected layers. The finished trained model was evaluated on a separate test set, where the model performed reasonably well. Additionally, we used the Adience benchmark, where our model, unfortunately, proved inferior to other models. We believe, however, that the latter result greatly underestimates the performance of our model.

Looking at future work, several improvements can increase the performance of our model. Firstly, the model should be trained with the same age-bins and evaluated on the Adience benchmark to obtain a better indication of the performance of our model. Secondly, the data that was used to train our model is quite uneven. Obtaining data for the bins that are not well represented in the data could facilitate for a reduction in the model's bias, as well as increase the model's performance. Furthermore, more data in terms of samples will most likely improve the performance of our model. Lastly, transfer learning can greatly improve the performance of our model. By using a model that is pre-trained on a large number of facial images, the model can be imported and fine-tuned, which probably would result in a better performing model.

There are, in addition, several methods that can be utilized to increase the performance of our model in the future. One is the concept of self-attention, outlined by \cite{waswani2017}, where CNN architectures can be adapted to focus more on fine-grained features of age-sensitive areas \citep{zhang2019finegrained}.  
Another approach could be by using Gabor wavelets in combination with a CNN to help the model extract global and local features, such as the work of \cite{gabor_wavelets}. 

With respect to the practical applicability of our model, there are several key points that have to be taken into consideration. As of now, it is quite easy to fool our model, either by holding up an image of another person or by performing grimaces to create wrinkles - fooling the model to predict an older age. This is something that has to be dealt with before using our model on real world problems. Nevertheless, as the world becomes more data-driven than ever before, we believe that with increasing computing power, and more sharing of knowledge through the open-source community, age-estimation models can become increasingly accurate and widespread.

\newpage
\bibliography{references}

\begin{thebibliography}{26}
\providecommand{\natexlab}[1]{#1}
\providecommand{\url}[1]{\texttt{#1}}
\expandafter\ifx\csname urlstyle\endcsname\relax
  \providecommand{\doi}[1]{doi: #1}\else
  \providecommand{\doi}{doi: \begingroup \urlstyle{rm}\Url}\fi

\bibitem[Abadi et~al.(2016)Abadi, Agarwal, Barham, Brevdo, Chen, Citro,
  Corrado, Davis, Dean, Devin, et~al.]{tensorflow}
Mart{\'\i}n Abadi, Ashish Agarwal, Paul Barham, Eugene Brevdo, Zhifeng Chen,
  Craig Citro, Greg~S Corrado, Andy Davis, Jeffrey Dean, Matthieu Devin, et~al.
\newblock Tensorflow: Large-scale machine learning on heterogeneous distributed
  systems.
\newblock \emph{arXiv preprint arXiv:1603.04467}, 2016.

\bibitem[Chao et~al.(2013)Chao, Liu, and Ding]{chao_et_al}
W.L. Chao, J.Z. Liu, and J.J. Ding.
\newblock Facial age estimation based on label-sensitive learning and
  age-oriented regression.
\newblock \emph{Pattern Recognition}, 46:\penalty0 628--641, 2013.
\newblock \doi{https://doi.org/10.1016/j.patcog.2012.09.011.}

\bibitem[Clark(2020)]{PIL}
A.~Clark.
\newblock \url{https://pillow.readthedocs.io/en/stable/}, 2020.

\bibitem[Dornaika et~al.(2020)Dornaika, Bekhouche, and
  Arganda-Carreras]{dornaika_et_al}
F.~Dornaika, S.~E. Bekhouche, and I.~Arganda-Carreras.
\newblock Robust regression with deep cnns for facial age estimation: An
  empirical study.
\newblock \emph{Expert Systems with Applications}, 141:\penalty0 1, Mar 01
  2020.
\newblock URL
  \url{https://search.proquest.com/docview/2319471430?accountid=12870}.

\bibitem[{Eidinger} et~al.(2014){Eidinger}, {Enbar}, and
  {Hassner}]{adience_authors}
E.~{Eidinger}, R.~{Enbar}, and T.~{Hassner}.
\newblock Age and gender estimation of unfiltered faces.
\newblock \emph{IEEE Transactions on Information Forensics and Security},
  9\penalty0 (12):\penalty0 2170--2179, 2014.
\newblock \doi{10.1109/TIFS.2014.2359646}.

\bibitem[Geitgey(2020)]{face_recognition}
A.~Geitgey.
\newblock \url{https://github.com/ageitgey/face_recognition}, 2020.

\bibitem[Goodfellow et~al.(2016)Goodfellow, Bengio, and
  Courville]{goodfellow_et_al}
I.~Goodfellow, Y.~Bengio, and A.~Courville.
\newblock \emph{Deep Learning}.
\newblock The MIT Press, 2016.
\newblock ISBN 0262035618.

\bibitem[GSB(2020)]{ntnu_bucket}
GSB.
\newblock Finished preprocessed data in google storage bucket.
\newblock
  \url{https://console.cloud.google.com/storage/browser/ntnu-ml-bucket?project=banded-splicer-294510&pageState=(\%22StorageObjectListTable\%22:(\%22f\%22:\%22\%255B\%255D\%22))&prefix=&forceOnObjectsSortingFiltering=false},
  2020.

\bibitem[Gu et~al.(2018)Gu, Wang, Kuen, Ma, Shahroudy, Shuai, Liu, Wang, Wang,
  Cai, and Chen]{gu_et_al}
J.~Gu, Z.~Wang, J.~Kuen, L.~Ma, A.~Shahroudy, B.~Shuai, T.~Liu, X.~Wang,
  G.~Wang, J.~Cai, and T.~Chen.
\newblock Recent advances in convolutional neural networks.
\newblock \emph{Pattern Recognition}, 77:\penalty0 354--377, 2018.

\bibitem[Kandel et~al.(2020)Kandel, Castelli, and Popovič]{optimizers_et_al}
I.~Kandel, M.~Castelli, and A.~Popovič.
\newblock Comparative study of first order optimizers for image classification
  using convolutional neural networks on histopathology images.
\newblock \emph{Journal of imaging}, 6\penalty0 (9):\penalty0 92, 2020.

\bibitem[King(2009)]{dlib}
D.~E. King.
\newblock Dlib-ml: A machine learning toolkit.
\newblock \emph{Journal of Machine Learning Research}, 10:\penalty0 1755--1758,
  2009.

\bibitem[Kingma and Ba(2017)]{adam_optimizer}
D.P. Kingma and J.~Ba.
\newblock Adam: A method for stochastic optimization, 2017.

\bibitem[Krizhevsky et~al.(2012)Krizhevsky, Sutskever, and Hinton]{alex_net}
A.~Krizhevsky, I.~Sutskever, and G.~E. Hinton.
\newblock Imagenet classification with deep convolutional neural networks.
\newblock In \emph{Proceedings of the 25th International Conference on Neural
  Information Processing Systems}, volume~1, page 1097–1105, Red Hook, NY,
  USA, 2012. Curran Associates Inc.

\bibitem[{Kwon} et~al.(2019){Kwon}, {Il Koo}, {Soh}, and {Ik
  Cho}]{gabor_wavelets}
H.~J. {Kwon}, H.~{Il Koo}, J.~W. {Soh}, and N.~{Ik Cho}.
\newblock Age estimation using trainable gabor wavelet layers in a
  convolutional neural network.
\newblock In \emph{2019 IEEE International Conference on Image Processing
  (ICIP)}, pages 3626--3630, 2019.
\newblock \doi{10.1109/ICIP.2019.8803442}.

\bibitem[{Lecun} et~al.(1998){Lecun}, {Bottou}, {Bengio}, and
  {Haffner}]{le_net}
Y.~{Lecun}, L.~{Bottou}, Y.~{Bengio}, and P.~{Haffner}.
\newblock Gradient-based learning applied to document recognition.
\newblock \emph{Proceedings of the IEEE}, 86\penalty0 (11):\penalty0
  2278--2324, 1998.
\newblock \doi{10.1109/5.726791}.

\bibitem[Phyo-Kyaw et~al.(2015)Phyo-Kyaw, Jian-Gang, and Eam-Khwang]{sai_et_al}
Sai Phyo-Kyaw, Wang Jian-Gang, and Teoh Eam-Khwang.
\newblock Facial age range estimation with extreme learning machines.
\newblock \emph{Neurocomputing}, 149:\penalty0 364--372, 2015.
\newblock \doi{10.1016/j.neucom.2014.03.074}.

\bibitem[Pontes et~al.(2016)Pontes, Britto, Fookes, and Koerich]{pontes_et_al}
J.~K. Pontes, A.~S. Britto, C.~Fookes, and A.~L. Koerich.
\newblock A flexible hierarchical approach for facial age estimation based on
  multiple features.
\newblock \emph{Pattern Recognition}, 54:\penalty0 34--51, 2016.

\bibitem[Prilianti et~al.(2019)Prilianti, Brotosudarmo, Anam, and
  Suryanto]{prillianti_et_al}
K.R. Prilianti, T.H.P. Brotosudarmo, S.~Anam, and A.~Suryanto.
\newblock Performance comparison of the convolutional neural network optimizer
  for photosynthetic pigments prediction on plant digital image.
\newblock \emph{AIP Publishing: University Park}, 2019.

\bibitem[Rothe et~al.(2018)Rothe, Timofte, and Van~Gool]{IMDB_dataset}
R.~Rothe, R.~Timofte, and L.~Van~Gool.
\newblock Deep expectation of real and apparent age from a single image without
  facial landmarks.
\newblock \emph{International Journal of Computer Vision}, 126\penalty0
  (2-4):\penalty0 144--157, 2018.

\bibitem[Smith and Chen(2018)]{mislabellings}
P.~Smith and C.~Chen.
\newblock Transfer learning with deep cnns for gender recognition and age
  estimation.
\newblock In \emph{2018 IEEE International Conference on Big Data (Big Data)},
  pages 2564--2571, 2018.
\newblock \doi{10.1109/BigData.2018.8621891}.

\bibitem[{Taigman} et~al.(2014){Taigman}, {Yang}, {Ranzato}, and
  {Wolf}]{deepface}
Y.~{Taigman}, M.~{Yang}, M.~{Ranzato}, and L.~{Wolf}.
\newblock Deepface: Closing the gap to human-level performance in face
  verification.
\newblock \emph{IEEE Conference on Computer Vision and Pattern Recognition},
  46:\penalty0 1701--1708, 2014.
\newblock \doi{10.1109/CVPR.2014.220}.

\bibitem[{Thukral} et~al.(2012){Thukral}, {Mitra}, and
  {Chellappa}]{hierarchical}
P.~{Thukral}, K.~{Mitra}, and R.~{Chellappa}.
\newblock A hierarchical approach for human age estimatio.
\newblock \emph{2012 IEEE International Conference on Acoustics, Speech and
  Signal Processing (ICASSP)}, pages 1529--1532, 2012.
\newblock \doi{10.1109/ICASSP.2012.6288182}.

\bibitem[{University of Massachusets}(2020)]{LFW}
{University of Massachusets}.
\newblock \url{http://vis-www.cs.umass.edu/lfw/results.html}, 2020.

\bibitem[Vaswani et~al.(2018)Vaswani, Shazeer, N., Uszkoreit, Jones, Gomez,
  Kaiser, and Polosukhin]{waswani2017}
A.~Vaswani, N.~Shazeer, Parmar N., J.~Uszkoreit, J.~Jones, A.~N. Gomez,
  L.~Kaiser, and I.~Polosukhin.
\newblock Attention is all you need, 2018.

\bibitem[Zhang et~al.(2020)Zhang, Lipton, Li, and Smola]{zhang2020dive}
Aston Zhang, Zachary~C. Lipton, Mu~Li, and Alexander~J. Smola.
\newblock \emph{Dive into Deep Learning}.
\newblock GitHub.com, 2020.
\newblock \url{https://d2l.ai}.

\bibitem[Zhang et~al.(2019)Zhang, Liu, Yuan, Guo, Gao, Zhao, and
  Ma]{zhang2019finegrained}
K.~Zhang, N.~Liu, X.~Yuan, X.~Guo, C.~Gao, Z.~Zhao, and Z.~Ma.
\newblock Fine-grained age estimation in the wild with attention lstm networks,
  2019.

\end{thebibliography}

\addappendix
\subsection{Adaptive Moment Estimation - Adam}
\label{app:adam}
Adam (Adaptive Moment Estimation), created by \cite{adam_optimizer}, is an optimizer which is based on the adaptive estimation of first-order and second-order moments. As outlined in \cite{optimizers_et_al}, the weights can be updated by the following equations: 
\begin{equation}
    w_t^i = w_{t-1}^i - \frac{\eta}{\sqrt{\hat{v^t}+\epsilon}} \hat{m_t} 
\end{equation}
where:
\begin{equation}
    \hat{m_t} = \frac{m_t}{1-\beta_1^t}
\end{equation}
\begin{equation}
    \hat{v_t} = \frac{v_t}{1-\beta_2^t}
\end{equation}

Here, $\eta$ is the learning rate, $w_t$ are the weights at step $t$, $m_t$ is the running average of the gradients (first moment), $v_t$ is the running average of the squared gradients (second moment). Furthermore, $\epsilon$ is a small constant for numerical stability, and $\beta_1^t$ and $\beta_2^t$ are used to select the amount of information needed from the previous updates, $\beta_i \in [0,1]$. All in all, the hyperparameters that is manually set before training are the learning rate $\eta$, epsilon $\eta$, $\beta_1$, and $\beta_2$.

\subsection{Overview of GitHub repository}
\label{app:github_repo}

Link to GitHub repository:  \href{https://github.com/christianbv/AgeNet}{GitHub repository}.

\subsection{Link to YouTube video}
\label{app:youtube_video}

Link to YouTube video:  
\href{https://www.youtube.com/watch?v=haNNlZm7L2o&ab\_channel=ChrisBv}{YouTube video}.


\end{document}